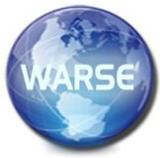

# Development of a Prototype Application for Rice Disease Detection Using Convolutional Neural Networks

**Harold Costales[1], Arpee Callejo-Arruejo[2], Noel Rafanan[3]**
[1] University of Northern Philippines, Philippines, hlcostales@unp.edu.ph
[2] University of Northern Philippines, Philippines, arpee.callejo@unp.edu.ph
[3] University of Northern Philippines, Philippines, noel.rafanan@unp.edu.ph

**ABSTRACT**

Rice is the number one staple food in the country, as this serves as the primary livelihood for thousands of Filipino households. However, as the tradition continues, farmers are not familiar with the different types of rice leaf diseases that might compromise the entire rice crop. The need to address the common bacterial leaf blight in rice is a serious disease that can lead to reduced yields and even crop loss of up to 75%. This paper is a design and development of a rice leaf disease detection mobile application prototype using an algorithm used for image analysis. The researchers also used the Rice Disease Image Dataset by Huy Minh Do available at https://www.kaggle.com/ to train state-of-the-art convolutional neural networks using transfer learning. Moreover, we used image augmentation to increase the number of image samples and the accuracy of the neural networks as well.

**Key words :** deep neural networks, convolutional neural networks, agriculture, transfer learning, rice diseases

## 1 INTRODUCTION

Across the 20th century, our prospects of engineering have rapidly changed. It was once thought of as impossible for everyone, as all we had seen on earth showed us a manual process that is uneasy for us; we would recognize as technology. However, in the 21st century, our machines continue to improve in what we conceive "Artificial Intelligence," which makes it more even feasible for devices to learn from experience, adjust to a high new level input and perform human-like tasks. Methods such as the use of convolutional neural networks may hold the key to developing software applications, particularly in the field of agriculture.

Portability, efficiency, and affordability of agricultural technology and information continue to be a major interference for improving agricultural productivity among small enterprises in the country. The Department of Agriculture, in partnership with the Department of Information and Communications Technology (DICT), has furnished a possible solution to enhance this kind of situation. Recently, they have launched the HACKATON to address further innovations that integrate software applications and systems into the development of agricultural technology, especially to farmers, which have been conducted nationwide. They have currently established outstanding software applications and systems to aid the farmers for better understanding when it comes to e-agriculture.

This paper emphasizes the role of ICT and the functional benefaction of software development to agriculture in the Philippines. Data from Rice Disease Image Dataset by Huy Minh Do available at https://www.kaggle.com are used to train data using a convolutional neural network algorithm using transfer learning. Moreover, we used image augmentation to increase the number of image samples and the neural network accuracy as well. It turns out that the modified neural networks achieved a state-of-the-art result with an accuracy closed to human-level performance. The researchers have developed a prototype application for Rice Leaf Disease Detection using Convolutional Neural Networks (CNNs) for farmers in the local community. This application will primarily install and use the said application on their smartphone, simply took a picture of the infected area of the rice leaf. Then the app gives a percentage of accuracy of rice disease infected in rice leaf.

Farmers were also able to grasp the knowledge of the different rice leaf diseases because of the information that the application is providing. This paper recommends the adoption of such software applications by institutions such as the Department of Agriculture to improve provision for appropriate decision making by agricultural farmers in the country.

Related studies on rice leaf disease detection using neural networks have been on-trend. According to [1] , "the use of computational intelligence-based techniques has proven successful in recent times for automated rice-disease detection (p.21)". Another related study stated that [3] "an automated system could have a feature on detection of diseases present in a rice leaf using color image analysis". Furthermore, [4] cited that "the management of perennial fruit crops requires close monitoring especially for the management of diseases that can affect production significantly and subsequently the post-harvest life (p. 1)". Corollary to the contexts presented, the researchers came up with a rice disease detection that integrates the use of an algorithm, specifically CNNs and an image analysis, for immediate monitoring of the rice crops.

Thus, this mobile application is needed to bring virtual IT experts into the field to determine, examine, and to give accurate results (on rice leaf diseases) that would inform the farmers what to do next without any further expenses.

### 1.1 Objectives of the Study

This study aimed to design and develop a mobile application for rice disease detection. Specifically, it sought to answer the following objectives:

1. Propose an application to address the problem, issues and challenges encountered;

2. Identify an appropriate algorithm for the application; and

3. Features for the Mobile application.

### 1.2 Conceptual Framework

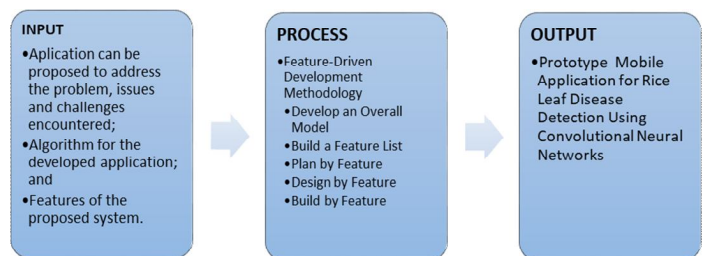

**Figure 1:** Paradigm of the Study

Figure 1 shows researcher used the input-process-output (IPO) model as a guide in conducting the study. The Application is proposed for the problems and issues encountered, appropriate algorithm to be incorporated, and features for the mobile application are the input of the study (input), which served as the guide in designing the application. The software development methodology, which is the Feature-Driven Development Methodology, is the process for the "Development of a Prototype Application for Rice Disease Detection Using Convolutional Neural Networks"(output).

## 2 METHODS

This part presents the algorithm and the software development methodology used in the development of the mobile application.

For the algorithm, the researchers chose Convolutional Neural Networks (CNNs) for the mobile application. As stated by [12], technological advancements in Computer Vision and Deep Learning a subset of the Artificial Intelligence gains more importance in the last decade especially in the field of object detection using Convolutional Neural Networks (CNN) became popular in many fields to address the current societal issues.

### 2.1 Convolutional Neural Networks Deep Learning Algorithm for the Prototype Mobile App

A Convolutional Neural Network (CNN) is a deep learning algorithm in which images serve as input to a learnable process to analyze various aspects of the image and be able to differentiate one from the other. A rice leaf is an input, and images of diseases of rice leaves are stored in the database to match the input. As cited by [11], many researchers use deep learning method to classify an image automatically. The purpose of classification is to arrange objects that will be observed into categories that have been defined. Furthermore, [14] cited that the detection technique assisted by simple image processing in the evidence is very interesting to be further researched. The researcher also utilized the method of [16] for data processing for the realization of the study.

Convolutional Neural Networks (CNNs) is the most appropriate model of Rice Disease Image Dataset since it uses image analysis. It was proven by [6] that convolutional neural networks (CNNs) have used in the field of computer vision for decades. Moreover, [5] proposes a convolutional autoencoder deep learning framework to support unsupervised image features learning for lung nodule through unlabeled data, which only needs a small amount of labeled data for efficient feature learning. Below are the steps on how the researchers came up with the integration of the CNNs for the developed application following the process of the CNNs in the study of [6].

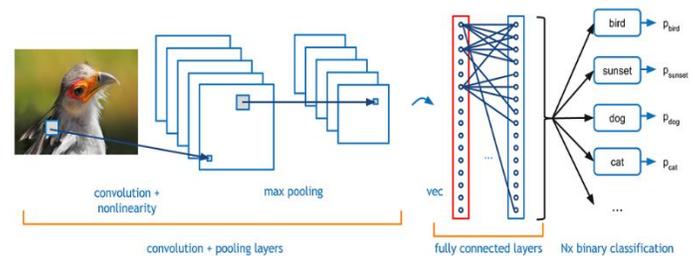

**Figure 2**. A Typical Architecture of CNNs by Tajbakhsh (2016)

Figure 2 presents a typical architecture of a CNN. It consists of an input layer (images), convolutional layer, pooling layer, a fully connected layer, a classifier and an output layer.

The convolutional layer is the most important layer in the CNN hence the name Convolutional Neural Networks. It acts as an automatic feature that extracts meaningful features of the object in the image. In mathematics, convolution is the operation of two functions to produce a third modified function.

$$(f * g)(t) \triangleq \int_{-\infty}^{\infty} f(\tau)g(t-\tau)\,d\tau.$$

(1)

In the context of CNNs, convolutional operation is simply a matrix operation, specifically matrix multiplication and addition. The byproduct of this operation called the feature map.

The pooling layer is used to down sample the feature map to reduce computational complexity. It is common in the literature to add the pooling layer after the convolution layer. There are two main types of pooling the max pooling and the average pooling. In the literature max pooling is preferred. It calculates the maximum value for each patch of the feature map.

$$M_f(p) = \max_{q \in \mathcal{G}}(f(q) - d_{\max}(p-q))$$

(2)

The fully connected layer, abbreviated FC is responsible for vectorizing the features from the convolution and pooling layers. This transforms the multi-dimensional feature map into a row vector to be fed into a classifier.

The classifier is responsible in outputing the probibilities of the output layer. There are two commonly used activation functions used in classifiers, the sigmoid activation function and the softmax activation function. Since the problem of detecting disease on rice leaf images is a multi-class classification problem, the appropriate function to be used is the softmax activation function.

$$\sigma(x_j) = \frac{e^{x_j}}{\sum_i e^{x_i}}$$

(3)

Another is the use of a software development methodology which helped the researchers identify the problems, created an overview of the concept for the mobile application and built the application.

## 2.2 Software Development Methodology

The researcher used the Feature-driven Development (FDD) under the family of Agile methodology for the software development because it is the most suitable and it has a customer-centric process. Its iterative feature allowed the researchers to develop the application while it is being tested. In the study of [7] stated that, "the iterative feature of the methodology allowed the proponents to capitalize on the learning that was accumulated during the development of earlier parts or versions of the solution". The figure below presents the phases of FDD methodology.

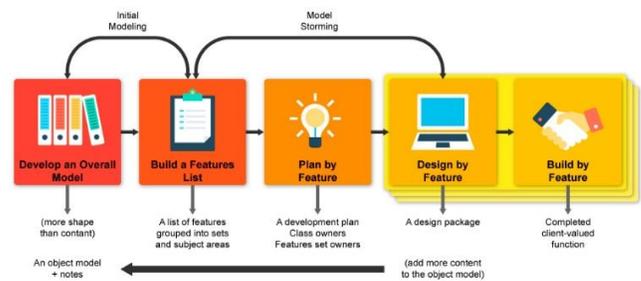

**Figure 3:** Feature-Driven Development Methodology

Figure 3 shows the Feature-Driven Development Methodology which has five phases, which include the (a) Development of an Overall Model, (b) Build a Feature List, (c) Plan by Feature, (d) Design by Feature, and (e) Build by Feature. The phases of development are presented as follows.

### 2.2.1 Develop Overall Model

A stage where the researchers created a fundamental foundation of the application and the variables it required. This stage also identified the initial development of the application. The researchers designed the application's overall model and served it as a guideline in developing the application. As the primary objective, the researchers develop the mobile app for the farmers for the early detection of rice diseases of rice-crops.

### 2.2.2 Build Feature List

The researchers created concrete plans for each feature of the mobile application. The researchers created a prototype of the user interfaces for the conceptualization of the mobile application.

### 2.2.3 Plan by Feature

The researchers created concrete plans for each feature of the mobile application. The researchers created a prototype of the user interfaces for the conceptualization of the mobile application.

### 2.2.4 Design by Feature

The researchers designed the features individually using the needed preferences. To fulfill this stage, the researchers designed the identified features with the help of the needed tools and processes.

### 2.2.5 Build by Feature

The last stage of the methodology wherein the researchers created the planned and designed features for the application. With the previous step as a guide, the researchers turned the details of the features into a working application, with the help of the tools and processes. After an inspection via a test run by the researchers, the researchers created the planned and designed features for the application.

## 3 RESULTS

### 3.1 Prototype Mobile Application for Rice Disease Detection

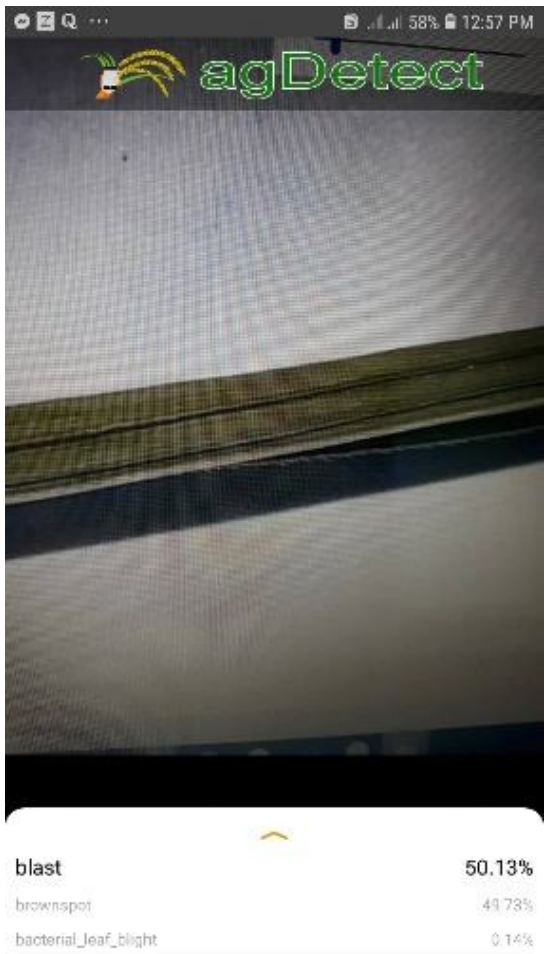

**Figure 4:** The Prototype of the Mobile Application

Figure 4 shows the rice disease detection of the mobile application. The researchers designed and develop a mobile application to assist the farmers in disease detection. It aims to help the farmers who are the backbone of the country to increase rice crop productivity. The lack of knowledge of farmers in the detection of rice diseases made the researchers conceptualize the integration of the field of agriculture and medicine in detecting diseases of rice crops. Also, Duterte's Administration supports Research and Development along with agriculture. It is stated in the Philippine Development Plan (PDP) 2017-2022 that there is a need to conduct researches in the field because of rice-crop yield losses. Thus, the PDP 2017-2022 established a plan to support the initiatives of the agriculture sector along with R&D.

Moreover, the application is different from other similar apps available in the market because it was developed specifically for the Philippine agriculture setting. The application will be stored with rice disease data exclusively for agriculture in the northern Philippines and further validated by pathologists as the researchers continue to gather data that will be for the database of the system.

### 3.2 Convolutional Neural Networks Deep Learning Algorithm for the Prototype Application

The researchers utilized an Image Processing typical machine learning workflow in the development of the model which consists of the following Data Acquisition, Data Preparation, and Training and Validation. As cited by [10] Image processing is the method of using different manipulation techniques and algorithms so that a desired features can be extracted such as morphology, color and texture from an image. Another study stated that [13] CNN proved to be best among the results in terms of accuracy when compared to other classifiers

#### 3.2.1. Image Acquisition

The dataset by Huy Minh Do available at https://www.kaggle.com/ containing 1,260 labeled rice leaf images was used for creating the model. The images were labeled as Leaf Blast, Brown Spot, and Hispa, which are the (3) three rice leaf diseases that the model will try to classify.

#### 3.2.2. Image Preprocessing

The images were downscaled to 500 x 500 pixels to reduce computational complexity. Image dataset were split into train dataset (80% of the entire samples) and validation dataset (20% of the entire samples). Image augmentation such as flipping, shearing, and rescaling in order to increase dataset samples as neural networks are data crunchers.

#### 3.2.3. Training and Validation

To fast track the training process the researchers employed transfer learning. Transfer learning, in general, is the process of taking a previously trained model used in a problem and apply in to another related problem. It is referring to the knowledge transfer from pretrained network in one domain to your own problem in a different domain. .[8]

The researcher used MobileNet 2 [9] as the base model. It is a state-of-the-art CNN model trained from ImageNet and one of the winners of the ImageNet Large Scale Visual Recognition Challenge (ILSVRC), a prestigious computer vision competition. Since the target dataset is small and somewhat different from the dataset where MobileNet 2 was trained, the top layer of the network needs to be frozen.

**First Iteration**
Base model mobile net Version 2.0
Hyper-parameters:
Top layer = false
Initialweights

**Second Iteration**
Base model mobile net Version 2.0
Hyper-parameters:
Top layer = false
Initialweights = imagenet
Loss = categorical cross entropy
Metrics = accuracy
Optimization = ADAM
EPOCHS= 10
TRAINABLE = FALSE
Result= 97% trained data set, 94% validation
Suffered from overfitting

**Third Iteration**
Base model mobile net Version 2.0,
image augmentation
Hyper-parameters:
Top layer = false
Initialweights = imagenet
Loss = categorical cross entropy
Metrics = accuracy
Optimization = ADAM
EPOCHS= 20
TRAINABLE = true
Result= 98.9% trained data set, 98% validation
Overfitting was solved

The third iteration proved that the process for the image analysis on the training and testing for the data sets was validated with an accuracy rate of 98% from 94% from the second iteration. This proved that the CNNs algorithm of the mobile application was proven for use.

**3.3 Features of the Prototype Mobile Application**

The features of the application include (1) real-time detection, (2) artificial intelligence, and (3) image analysis.

**3.3.1 Real-time Detection**

One of the features of the mobile application is a real-time detection. The application helps in the identification of the rice leaf diseases using convolutional neural networks that match the input data and the data stored in the database of the application.

The embed camera of the mobile application can detect the diseases in real-time. This feature helps farmers to detect the diseases even without the instruction of plant pathologists.

**3.3.2 Artificial Intelligence**

Experts in the field of Agriculture are marginal, especially in rice leaf disease detection. This feature of the mobile application replicates the knowledge about rice leaf disease detection in the field, which is one of the features of the mobile application. As data grows on the database of the mobile application, the more accurate results it can give.

**3.3.3 Image Analysis**

Another feature is Image Analysis. This feature uses the CNNs algorithm aspects and features of images were stored, enabling the mobile application to analyze the rice leaf for disease detection through the following steps of the algorithm.

**4 CONCLUSION**

In summary, the researchers have developed a mobile application, which could help the agricultural sector. Specifically, the researchers conclude that the mobile application will help in the realization of the high-yield of rice crops against rice diseases for preventive measures. Thus, this study is helpful to the country, particularly the agricultural sector.

**5 RECOMMENDATION**

The researchers came up with the following recommendations:
(1) The output of the Mobile Application will have its validation of experts in rice disease for accuracy of the information from the mobile application, (2) the mobile application will have a series of tests for the implementation for future use. Lastly, (3) the researchers highly recommend that the application shall be introduced to the Department of Agriculture for dissemination.

**ACKNOWLEDGEMENT**

The researchers would like to thank the following people: the active UNP Research Office- Science and Technology Coordinator, Prof. Redentor S. Rojas, the ever-supportive UNP Research Director, Dr. Edelyn Cadorna and the dynamic UNP President, Dr. Erwin F. Cadorna for pushing us to enhance our research knowledge and capabilities. Also, the researchers would like to thank the IMPACT HACKATON 2050 for organizing the IMPACT HACKATON 2019, where they conceptualized this research study. Above all, thanks to Almighty God for all the blessings for our families and the society.